%% file: main.tex
\documentclass[11pt]{article}

\usepackage[preprint]{acl}
\usepackage{times}
\usepackage{latexsym}
\usepackage[T1]{fontenc}
\usepackage[utf8]{inputenc}
\usepackage{microtype}
\usepackage{inconsolata}
\usepackage{graphicx}
\usepackage{booktabs}
\usepackage{array}
\usepackage{tabularx}
\usepackage{tikz}
\usetikzlibrary{arrows.meta}
\usepackage{pgfplots}
\pgfplotsset{compat=1.18}
\usepgfplotslibrary{groupplots}
\usepackage{amsmath,amssymb}

\renewcommand{\emph}[1]{#1}
\newcommand{\taur}{$\tau$-retail}

\usepackage{xspace}
\newcommand{\ie}{i.e.\@\xspace}

\title{Handoff Curves: Initial Selection, Later Handoff, and Supervision Targets\\
in Multi-turn LLM Routing}
\author{Daniel Matos \qquad Lu Liu \qquad
  Yanfeng Zhang\thanks{Corresponding author: \texttt{zhangyanfeng8@huawei.com}} \\ Riemann Lab, Huawei Technologies}

\begin{document}
\maketitle

\begin{abstract}
We study how initial model selection, later handoff, and training targets affect success and
cost in LLM tasks that span multiple turns. We compare several router architectures using
handoff curves, which record task success and cost as the handoff step changes. Our dataset
contains 34,204 complete executions of three Qwen models across BFCL, MultiWOZ, and \taur{},
providing offline supervision for routing decisions. In our primary comparison, with Qwen3.5 35B-A3B as the
starting model, allowing a later handoff does not establish an improvement in utility over
initial selection on BFCL or MultiWOZ, and increases cost on BFCL across a range of cost weights.
Changing the supervision targets improves router performance on both our dataset and
TwinRouterBench. On our dataset, replacing noisy training success curves with monotone fits
raises success by 0.2 to 4.1 percentage points at the same cost across the reported operating
points. On TwinRouterBench, we construct targets that describe the model capability needed for
the remaining work and difficult steps ahead. Compared with training on the benchmark's per-step labels,
these targets raise trajectory pass scores by 7.8 percentage points and cost savings by 6.3
percentage points on the full static bank. A live deployment of the router trained with these
targets resolves 79 of 100 instances at a router cost of \$0.59 per instance.
\end{abstract}

\input{sections/1_introduction}
\input{sections/2_related_work}
\input{sections/3_methodology}
\input{sections/4_datasets}
\input{sections/5_results}

\input{sections/6_discussion}

\bibliography{custom}

\appendix

\input{sections/a_prediction_live_policy}
\input{sections/b_execution_inventory}
\input{sections/c_evaluation_details}
\input{sections/d_secondary_comparisons}

\end{document}

%% file: sections/1_introduction.tex
\section{Introduction}
\label{sec:intro}

Large language model (LLM) routing decides which model should serve each query,
task, or user, usually trading answer quality against cost. Early LLM routing work used cost-aware
cascades~\cite{chen2023frugalgpt} and routers trained on benchmark
outcomes~\cite{shnitzer2023routing}, and later work added pairwise routers and shared
benchmarks~\cite{ding2024hybrid,ong2025routellm,hu2024routerbench}. These routers
were designed for a single query, but the field has since moved to multi-turn
routing~\cite{zhang2026mtrouter,zhang2026myopic,xu2026seqroute,yang2026twinrouterbench}.

Existing multi-turn routers construct supervision from realised trajectories,
verified per-step downgrades, search-derived policies, or annotated handoff
windows~\cite{zhang2026mtrouter,yang2026twinrouterbench,zhang2026myopic,lei2026tacit}. These labels use outcomes observed later in the interaction, but
they do not provide repeated outcomes for committing at each alternative handoff step
of the same task.
The Handoff Tax measures handoffs at selected switch points~\cite{ganz2026handoff}, and
the Replay Gap shows that substituting logged outputs does not reproduce a live
continuation~\cite{gonuguntla2026replay}. Both results motivate direct measurement of the
handoff curve.

We collect these curves on BFCL~\cite{patil2025berkeley}, \taur{}~\cite{yao2024taubench},
and MultiWOZ~\cite{budzianowski2018multiwoz} by committing to a model handoff at
each collected handoff step and measuring task success and cost. These measurements
let us ask how much of a router's success and cost trade-off comes from initial
selection, and whether allowing a later handoff adds value.

Each task begins on a starting model, which a router can keep or replace.
We evaluate routers from each starting model in our pool (Qwen3 8B, Qwen3.5 35B-A3B, and Qwen3.5 397B-A17B), covering both
upward and downward handoffs, to a larger or a smaller model, respectively. We compare two policies that use the same fitted
router. Initial selection chooses a model at the first step and keeps it.
Later handoff makes the same first choice and, if that choice keeps the
starting model, can then hand off to another model at a later step.
At the primary 35B start on BFCL and MultiWOZ, the direct router's later-handoff
policy adds no established benefit over initial selection. Initial selection by
the factored router improves utility over the training-selected fixed policy at
one tested cost weight on BFCL, but the fixed policy has higher utility at
$\lambda\in[0.2,2]$ on BFCL and is not beaten on MultiWOZ.
We also propose new supervision targets,
that is, what the router learns to predict. On our dataset, we make each training
success curve monotone across handoff steps, either increasing or decreasing depending on the type of handoff. On
TwinRouterBench~\cite{yang2026twinrouterbench}, which has no measured handoff curves, we
adapt this idea by replacing each step's tier label with the strongest tier
needed from that step to the end of the instance. On our dataset, these targets raise the direct router's
utility under initial selection at 7 of the 15 cost weights on BFCL and at 4 on
MultiWOZ, with no established decrease at any tested weight. They raise the success of both policies by
similar amounts, so later handoff still adds no established benefit over initial
selection. On TwinRouterBench, they improve the static scores.
Our contributions are as follows:
\begin{itemize}

\item We build a dataset of $34{,}204$ executions on BFCL,
\taur{}, and MultiWOZ, with replicated handoff curves, billed costs, and both
successful and failed outcomes, which we will release publicly.

\item We investigate whether a later handoff adds value over initial selection,
from every starting model and with the same fitted router.

\item We show that changing
the supervision targets improves routing at several cost weights on our dataset and in
static TwinRouterBench evaluation, under an identical training procedure. In the live
TwinRouterBench track, a router trained with these targets resolves 79 of 100 instances at a router cost of \$0.59 per instance.

\end{itemize}

%% file: sections/2_related_work.tex
\section{Related Work}
\label{sec:related}

\paragraph{Routing policies and supervision.}
Single-query routers and cascades send each query to one model or escalate it to a
stronger one~\cite{chen2023frugalgpt,shnitzer2023routing,ding2024hybrid,ong2025routellm,aggarwal2024automix}.
Multi-turn routers learn from trajectory returns or searched policies, as in
MTRouter~\cite{zhang2026mtrouter}, DialRouter~\cite{zhang2026myopic}, and
SeqRoute~\cite{xu2026seqroute}. Other methods replace or defer individual steps, as in
TRIM~\cite{kapoor2026trim}, DMR~\cite{li2026dynamic}, and ReDAct~\cite{piatrashyn2026redact},
or restart a task on a stronger model when it escalates, as in
SWE-Router~\cite{son2026swe}. These methods make other routing decisions.
TACIT-Switch learns when to make one permanent upward handoff, from one rollout per
model and a teacher-annotated handoff window~\cite{lei2026tacit}. We also study one
permanent handoff, but we measure its outcome at every collected handoff step of the same
task.

\paragraph{Handoff measurement and routing datasets.}
RouterBench provides model--query outcomes without a within-session timing
axis~\cite{hu2024routerbench}. TwinRouterBench verifies local tier substitutions (tiers
are model levels from 0, the cheapest, to 3, the strongest) by executing continuations,
retaining substitutions only when the episode passes~\cite{yang2026twinrouterbench}. The
Handoff Tax evaluates seven fixed switch points~\cite{ganz2026handoff}, the switch-drift study
examines a final-turn handoff~\cite{khraishi2026evaluating}, and the Replay Gap
distinguishes live continuations from logged-output substitution~\cite{gonuguntla2026replay}.
Our dataset measures successful and failed committed handoffs at collected handoff steps,
with two executions of each policy on each task, three starting models, and three
datasets. It supplies outcomes and billed costs for router supervision.

\paragraph{Supervision and routing decisions.}
Routing gains depend on the model pool and its shared failures, on task composition, and
on label noise~\cite{chen2026does,lee2026most}. The gap to oracle performance depends on
how the oracle is defined under stochastic decoding~\cite{chen2026much}, and deployable
routers realise only part of it~\cite{shihab2026opportunity}. Accurate outcome predictions also
need not produce better decisions~\cite{zhang2026calibration}. As in that work, we compare
supervision targets under an identical training procedure and decision rule. We also
compare initial selection with later handoff under the same fitted router.

%% file: sections/3_methodology.tex
\section{Methodology}
\label{sec:method}

\subsection{Measuring Handoff Outcomes}

For task $x$, starting model $a$, and receiving model $b$, the handoff policy
$\pi_k^{a\rightarrow b}$ lets $a$ take the first $k$ steps and $b$ take all remaining steps, with no further handoff. We call step $k$ the handoff step. The
handoff curve is the
success and cost of this handoff as functions of $k$:
\begin{equation}
\begin{aligned}
q_{a\rightarrow b}(x,k)
 &= \mathbb{E}_{\pi_k^{a\rightarrow b}}[\,Y\mid x\,],\\
c_{a\rightarrow b}(x,k)
 &= \mathbb{E}_{\pi_k^{a\rightarrow b}}[\,C\mid x\,].
\end{aligned}
\label{eq:quality-cost-curves}
\end{equation}
Here, $Y\in\{0,1\}$ is the binary task success, and $C$
is the billed cost of the execution.  The expectation is over executions of the
policy on task $x$. We execute every policy and do not estimate the curves from
logged trajectories. A solo execution of $b$ supplies the $k=0$ outcome, while
staying on $a$ (a solo execution of $a$) is a separate action. Each cell in the dataset is one policy $\pi_k^{a\rightarrow b}$ on task $x$, or
one solo model, and contains two separately collected end-to-end
executions. Each execution runs the whole episode, including the first $k$ steps by $a$ (the
prefix). The mean success of the two executions is the cell's success label, which
averages over randomness in model outputs. The executions share
the task and collection conditions, and are not assumed to be statistically
independent. 

\subsection{Supervision Targets}
\label{sec:method-supervision}

A supervision target is the quantity that a router learns to predict. We change
it while the learner, features, splits, and decision rule stay fixed,
see Figure~\ref{fig:supervision}. 

On our dataset, raw labels are the measured success probability of each cell. Projected
labels make each training success curve monotone across handoff steps by
isotonic regression~\cite{barlow1972isotonic}: nonincreasing for upward handoffs
and nondecreasing for downward handoffs. We expect this direction because the later
an upward handoff, the more steps the weaker model takes, and the later a downward
handoff, the more steps the stronger model takes.
With two executions per cell, the raw curves are noisy, so the projection replaces
each curve with the closest curve that has this direction. More executions per cell
would also reduce the noise, but each one costs a full episode. We compare raw and
projected labels for the direct router on BFCL and MultiWOZ, see
Table~\ref{tab:label-effects}. Only the training curves are projected. Both
versions of the router are scored on the measured, unprojected outcomes of the
evaluation tasks. The projection is a modelling choice for the training targets.
It does not claim that the true success curves are monotone.

TwinRouterBench~\cite{yang2026twinrouterbench} has no measured handoff curves.
Its static bank, see Section~\ref{sec:datasets-trb}, gives each step $j$ a benchmark label $r_j\in\{0,\ldots,3\}$: the cheapest tier
found by the benchmark's search for which step $j$ still passes, given the tiers already accepted for earlier steps
and the strongest tier for all later steps (larger values are stronger models).
Along an instance, these labels can rise and fall. We adapt our monotone targets
to this setting by transforming the labels rather than fitting them. Each step
$k$ receives the forward target
\begin{equation}
R_k=\max_{j=k,\ldots,T}r_j,
\label{eq:remaining-work}
\end{equation}
the strongest tier that step $k$ or any later step needs, where $T$ is the final
labelled step. If the labels from step $k$ on are 1, 3, 1, then $R_k=3$, although
step $k$ itself needs only tier~1. $R_k$ can only fall as $k$ grows, see Figure~\ref{fig:supervision}b. So the
indicator $R_k\le t$, which says whether a permanent handoff to tier $t$ at step
$k$ is sufficient under the benchmark labels, can only rise. This is the same
monotone direction that our fit gives to the success curve of a downward handoff. We build forward targets from complete training instances and predict them from the observed prefix. These two dataset settings share the monotone direction of the targets but not the method: on our dataset,
a fit smooths measured outcomes, whereas on TwinRouterBench, a maximum transforms the original benchmark labels.

\input{sections/fig_supervision}

\subsection{Routers}
\label{sec:method-routers}

We study the direct, factored, and burst-aware routers, with SR-KNN as a baseline.
All four see only the prefix observed at the decision step
and embed text with a frozen MiniLM encoder~\cite{wang2020minilm,reimers2019sentence}.
Table~\ref{tab:routers} lists the data, supervision target, and model of each
router. The direct and factored routers run on our dataset and use the decision
rule in Section~\ref{sec:method-policies}. The burst-aware router runs on
TwinRouterBench and makes at most one permanent downward handoff.

\begin{table}[t]
\centering\small
\setlength{\tabcolsep}{4pt}
\begin{tabularx}{\columnwidth}{@{}ll>{\raggedright\arraybackslash}X>{\raggedright\arraybackslash}p{1.9cm}@{}}
\toprule
Router & Data & Supervision target & Model \\
\midrule
Direct & Ours & Raw or projected & MLP regressor \\
Factored & Ours & Projected (timing), raw (difficulty) & Two MLP regressors \\
Burst-aware & TRB & Benchmark or forward & Boosted-tree classifiers \\
SR-KNN & Both & Raw (ours), benchmark (TRB) & $k$-NN ($k=5$) \\
\bottomrule
\end{tabularx}
\caption{Routers, the data they run on, what they learn to predict, and their
model, see Section~\ref{sec:method-supervision} for the supervision targets. Ours: our dataset. TRB:
TwinRouterBench.}
\label{tab:routers}
\end{table}

\paragraph{Direct router.}
The direct router regresses the success probability of each cell with a
multilayer perceptron with hidden layers of 64 and 32 units and a sigmoid output,
trained with cross-entropy against the cell's success label. It reads the
instruction, the observed history, the handoff step, and the starting and
receiving models through frozen MiniLM
embeddings~\cite{wang2020minilm,reimers2019sentence}, 11 history statistics, such as
counts of actions and errors, and model, step, and price features. On BFCL, we
train it with ten task-grouped folds and six seeds, and on MultiWOZ and \taur{},
with fixed training and evaluation splits and five seeds, see
Appendix~\ref{app:codex-instruments}. Unless stated otherwise, results use the direct
router trained on raw labels.

\paragraph{Factored router.}
The factored router splits the same prediction into two parts. The first part,
$d$, predicts how likely the stronger model of the pair is to solve the task on
its own. It measures task difficulty and does not depend on timing. The second
part, $g(k)=\min\{\widetilde q(k)/d,1\}$, predicts how much of that success
remains when the handoff happens at step $k$, and is learned from the isotonic
projection $\widetilde q$ of each training curve, see
Section~\ref{sec:method-supervision}. The router scores each handoff by the
product $d\,g(k)$ of two regression models of the same kind as the direct router,
with $g$ trained on targets between 0 and 1. Unlike the direct router, which can
learn any pattern across handoff steps, the factored router separates difficulty
from timing.

\paragraph{Burst-aware router.}
The burst-aware router runs on TwinRouterBench, whose tiers range from 0, the
cheapest, to 3, the strongest. At each step, it either retains tier 3 or commits
to a cheaper tier $t\in\{0,1,2\}$ for the rest of the instance. A burst is a step
labelled $r_k=3$. Under the benchmark's scoring rule, a permanent handoff before
a later burst causes the instance to fail.

Three gradient-boosted tree classifiers~\cite{friedman2001gbm,ke2017lightgbm} read
the same features of the observed prefix, see Appendix~\ref{app:codex-instruments},
and each contributes one factor to the score. The tier-requirement classifier
predicts $R_k$, the tier needed from step $k$ onward, when trained on forward
targets. With benchmark labels, it predicts $r_k$, the tier needed at step $k$. In
either case, $p_{\mathrm{req}}(t)$ is the predicted probability that the required
tier is at most $t$. The burst classifier gives
$p_{\mathrm{clear}}$, the probability that step $k$ is not a burst. With forward
targets, it is instead the probability that neither step $k$ nor any of the next
three labelled steps is a burst. The local-feasibility classifier is trained only
on steps that are not bursts and gives $p_{\mathrm{loc}}(t)$, the probability that
$r_k\le t$ on such a step, which equals 1 for $t=2$. The router combines the
three outputs and calibrates the resulting score:
\begin{equation}
p_t=I_t\!\left(\sqrt{p_{\mathrm{req}}(t)\,p_{\mathrm{clear}}\,p_{\mathrm{loc}}(t)}\right),
\label{eq:tier-score}
\end{equation}
where $I_t$ is an isotonic regression~\cite{zadrozny2002transforming}
fitted entirely within the training data. We obtain its input scores through
three-fold cross-validation, keeping all steps of each instance together.
Each score therefore comes from classifiers trained without that instance.
Evaluation instances are excluded from both classifier fitting and calibration. With forward targets, the tier-requirement
classifier predicts sufficiency for the remaining trajectory, while the other
two classifiers assess local feasibility and the absence of a near-term burst.
Calibration targets $R_k\le t$ with forward targets and $r_k\le t$ with benchmark
labels. Thus the two versions use probabilities of different events to make
the same permanent-handoff decision.

A tier is eligible when $p_t\ge p_{\min}$. Among eligible tiers, the router
maximises the handoff score
\begin{equation}
V_t=p_t\,S_t-(1-p_t)\,F_t,
\label{eq:handoff-value}
\end{equation}
where $S_t$ forecasts the saving from a successful handoff and $F_t$ assigns a
cost penalty to failure. The router hands off when the largest score is positive,
preferring the cheaper tier on ties. Otherwise, it retains tier 3 and reconsiders
at the next step.

Static evaluation uses price-based cost proxies. The saving $S_t$ uses the
input-price difference and forecast remaining prompt volume. The penalty $F_t$
uses the strongest tier's input price and forecast whole-instance prompt volume,
scaled by a weight $\beta$. Inner cross-validation selects $p_{\min}$ and $\beta$,
see Appendix~\ref{app:codex-instruments}. For Table~\ref{tab:trb-supervision}, we train one
version on benchmark labels and one on forward targets. They differ only in the
targets of the tier-requirement and burst classifiers and in the calibration event,
and both are scored on the same five folds that hold out whole instances.

For live evaluation, we refit the forward-target router on the full static bank
and use a fixed floor of $p_{\min}=0.95$, live cost forecasts, and a spending and
step guard. Appendix~\ref{app:codex-instruments} specifies these settings.

\paragraph{SR-KNN baseline.}
As a reference, we use SR-KNN, the nearest-neighbour~\cite{cover1967nearest}
baseline of semantic routing~\cite{liu2026vllm}. We use a held-out, five-neighbour
version, whereas TwinRouterBench reports SR-KNN as an in-sample one-neighbour upper
bound~\cite{yang2026twinrouterbench}. On our dataset,
it averages the measured success of the five most similar training cells of the
same ordered model pair, weighted by similarity, and uses the decision rule of
Section~\ref{sec:method-policies}. On TwinRouterBench, it compares the mean tier
label of the five most similar training steps with a threshold selected by
cross-validation inside the training folds. It makes this comparison at the first
step and keeps the chosen tier, the cheapest or the strongest, for the whole
instance.

\subsection{Initial Selection and Later Handoff}
\label{sec:method-policies}

On our three datasets, we compare initial selection and later handoff using the
same fitted router: the direct router, the factored router, or SR-KNN. At $k=0$,
before any model has acted, both policies choose among the three models. Initial
selection keeps that choice for the entire task. Later handoff also commits
immediately if the choice replaces the designated starting model. If the choice
retains the starting model, it may reconsider at later collected steps that the
trajectory reaches, before the median task length in the training data, see
Appendix~\ref{app:codex-evaluation}.

At a later decision, a handoff occurs only when another model has strictly
greater predicted utility than staying. The receiving model then finishes the
task. If no handoff occurs, the starting model finishes it. The rule estimates
no separate value for waiting. Both policies share the fitted predictor,
opening, cost forecasts, and tie rule, with staying preferred on exact ties.

We evaluate each policy by mean task success $Q$ and mean billed execution cost
$C$. Each task contributes the selected cell's average outcome over its two
executions, with router initialisation seeds averaged within task. Utility is
\begin{equation}
U_\lambda=Q-\lambda_{\mathrm{abs}}C,
\label{eq:utility}
\end{equation}
where $Q$ is a fraction and $\lambda_{\mathrm{abs}}=\lambda/\bar C_a$, with
$\bar C_a$ the starting model's mean training solo cost, computed within each
fold. Costs in this expression are in dollars per task, and tables report
$10^{-3}$\,\$ per task. Router computation is excluded. At $\lambda=1$, an
additional cost equal to 1\% of $\bar C_a$ offsets one percentage point of
success. We evaluate 15 dimensionless weights from 0 to 50. Equal weights
across starts or datasets need not represent equal prices in dollars.

Both policies choose by predicted utility. At history $h_k$, a candidate model
$b$ receives
\begin{equation}
\widehat U_\lambda(b\mid h_k,a)
 = \hat q(b\mid h_k,a)-\lambda_{\mathrm{abs}}\hat c(a,b,k),
\label{eq:sequential-action}
\end{equation}
where $\hat q$ predicts terminal task success from the observed prefix, and
$\hat c(a,b,k)$ forecasts total episode cost from training cells. For $b=a$, the
score assumes that the starting model finishes the task. For another model, it
assumes a handoff now followed by permanent use of that model.

We measure the paired difference, later handoff minus initial selection, in
success, cost, and utility over all eligible evaluation tasks. Decisions use one
observed trajectory per task and starting model, while outcomes come from the
separately collected execution cell selected by the policy. This assumes that
the cell's outcome applies after the observed prefix. We assess sensitivity to
this assumption by restricting evaluation to identical prefixes.

We also compare initial selection with a fixed policy:
for each cost weight, the solo model or fixed handoff schedule, a handoff to a given model at a given step, with the
highest mean training utility, applied to every evaluation task,
see Appendix~\ref{app:codex-evaluation}.

Intervals resample tasks (10{,}000 paired resamples, seeds averaged within task)
and condition on the fitted models. Primary comparisons are the direct router at
the 35B start on BFCL and MultiWOZ, with simultaneous 95\% bands over all weights
and metrics, adjusted across the two datasets. Secondary comparisons (other
starts, other routers, and the routers against the training-selected fixed policy, on BFCL and
MultiWOZ) use unadjusted
95\% intervals. All other comparisons are descriptive, including \taur{}, the
router comparison at initial selection, the supervision targets, and
TwinRouterBench. We call a difference established
when its interval excludes zero. An interval containing zero does not establish
equivalence.

%% file: sections/fig_supervision.tex
\begin{figure}[t]
\centering
\begin{tikzpicture}[font=\scriptsize]
\begin{groupplot}[
  group style={group size=2 by 1, horizontal sep=1.15cm},
  scale only axis, width=2.45cm, height=2.2cm,
  axis lines*=left, axis line style={black!70},
  tick align=outside, tick style={black!60}, major tick length=2pt,
  xtick={1,2,3,4}, xmin=0.6, xmax=4.4, xlabel={Handoff step $k$},
  ymajorgrids, major grid style={black!12},
  title style={font=\scriptsize\bfseries, align=center, yshift=-2pt},
  xlabel style={yshift=2pt}, ylabel style={yshift=-3pt, align=center},
  clip=false,
]
\nextgroupplot[title={(a) Our dataset:\\ projected label},
  ylabel={Success (\%)}, ymin=0, ymax=100, ytick={0,50,100}]
\addplot[black!45, dashed, thin] coordinates {(1,80) (2,50) (3,70) (4,30)};
\addplot[black!85, very thick, mark=*, mark size=1.9pt]
  coordinates {(1,80) (2,60) (3,60) (4,30)};
\addplot[only marks, mark=x, mark size=2.1pt, black!55, thick]
  coordinates {(1,80) (2,50) (3,70) (4,30)};
\node[black!60, anchor=north, font=\tiny] at (axis cs:2,45) {raw};
\node[black!85, anchor=south west, font=\tiny, inner sep=1pt] at (axis cs:1.5,73) {projected};
\nextgroupplot[title={(b) TwinRouterBench:\\ forward target},
  ylabel={Tier\\(higher is stronger)}, ymin=-0.35, ymax=3.35, ytick={0,1,2,3}]
\addplot[black!45, dashed, thin] coordinates {(1,1) (2,0) (3,3) (4,1)};
\addplot[black!85, very thick, mark=*, mark size=1.9pt]
  coordinates {(1,3) (2,3) (3,3) (4,1)};
\addplot[only marks, mark=x, mark size=2.1pt, black!55, thick]
  coordinates {(1,1) (2,0) (3,3) (4,1)};
\node[black!60, anchor=west, font=\tiny] at (axis cs:2.15,0.0) {benchmark $r_j$};
\node[black!85, anchor=north, font=\tiny] at (axis cs:1.55,2.82) {forward $R_k$};
\end{groupplot}
\end{tikzpicture}
\caption{Supervision targets (illustrative values). (a) Raw success of an upward
handoff and its projected label. (b) Benchmark labels $r_j$ and forward targets
$R_k=\max_{j\ge k} r_j$.}
\label{fig:supervision}
\end{figure}
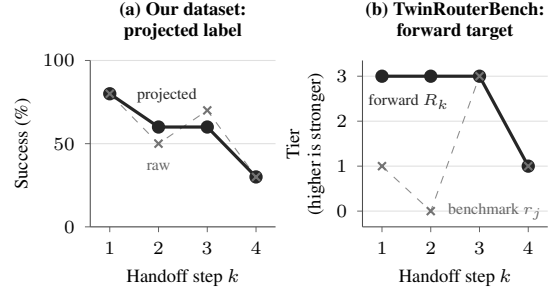

%% file: sections/4_datasets.tex
\section{Datasets}
\label{sec:datasets}

We use two complementary resources: our dataset measures task outcomes under
committed handoffs, while TwinRouterBench supplies verified benchmark labels for
comparing supervision targets, see Section~\ref{sec:results-supervision}.

\subsection{Source Datasets and Collection}

We collect handoff curves on three datasets. BFCL's \texttt{multi\_turn\_base}
category~\cite{patil2025berkeley} tests tool use across user turns, with decisions indexed by user turn. Its official metric checks the
environment state and the execution responses after every turn. We instead count a task
as solved when the model reaches the reference final state by any sequence of calls,
after normalising whitespace and trailing punctuation, and we exclude tasks whose
reference calls fail. \taur{}~\cite{yao2024taubench} pairs a retail agent with a
simulated user. Its decisions are indexed by agent action, and success requires
a final reward of one. MultiWOZ 2.4~\cite{budzianowski2018multiwoz,ye2022multiwoz} provides
dialogue state tracking tasks indexed by dialogue turn. A dialogue succeeds when
the F1 score between its final predicted state and the gold state, computed over
slot--value pairs, is at least $0.9$. This measures state tracking, not general
dialogue completion. On all three
datasets, task success is the share of executions that succeed. Because the
success criteria differ, equal success rates on two datasets do not mean equal
difficulty.

Collection begins with solo screening to identify tasks on which models differ.
Selected tasks receive handoff executions at the collected steps in
Table~\ref{tab:handoff-steps}, retaining both successful and failed outcomes. This screening
enriches routing opportunities, so the resulting populations are not
representative samples of the full benchmarks.

\begin{table}[t]
\centering\small
\setlength{\tabcolsep}{3pt}
\begin{tabularx}{\columnwidth}{@{}lcc>{\raggedright\arraybackslash}X@{}}
\toprule
Dataset & Upward & Downward & Step unit \\
\midrule
BFCL & 1--6 & 1--6 & User turn \\
MultiWOZ & 1--11 & 1--11 & Dialogue turn \\
\taur{} & 1, 2, 4, 6, 8 & 2, 4, 6, 8, 10 & Agent action \\
\bottomrule
\end{tabularx}
\caption{Collected handoff steps by dataset and direction. At step $k$, the
starting model has completed $k$ steps and the receiving model takes over.
Initial selection at $k=0$ uses solo executions. Evaluation considers reachable
collected steps before the training-derived cutoff described in
Appendix~\ref{app:evaluation-populations}.}
\label{tab:handoff-steps}
\end{table}

The model pool comprises Qwen3 8B~\cite{yang2025qwen3}, Qwen3.5 35B-A3B, and Qwen3.5 397B-A17B~\cite{qwen2026qwen35}, served
through OpenRouter. The \taur{} user simulator is GPT-4.1-mini~\cite{openai2025gpt41}, and BFCL and
MultiWOZ use scripted user turns. Dataset costs are agent-side execution costs, from provider bills except for 431
re-collected executions priced at list price.
Simulator spending is recorded separately where available and is not included in
the routing cost. Appendix~\ref{app:codex-data} gives further information.

\begin{table}[t]
\centering\small
\setlength{\tabcolsep}{3pt}
\begin{tabular}{@{}lrr*{3}{>{\raggedleft\arraybackslash}p{0.75cm}}@{}}
\toprule
& & & \multicolumn{3}{c}{Evaluation tasks} \\
\cmidrule(l){4-6}
Dataset & Cells & Executions & 8B & 35B & 397B \\
\midrule
BFCL & 2{,}247 & 4{,}494 & 102 & 102 & 102 \\
MultiWOZ & 12{,}504 & 25{,}008 & 117 & 117 & 117 \\
\taur{} & 2{,}351 & 4{,}702 & 41 & 17 & 17 \\
\midrule
Total & 17{,}102 & 34{,}204 & & & \\
\bottomrule
\end{tabular}
\caption{Dataset size. Cells and executions cover all tasks (training,
screening, and evaluation). The last three columns give the evaluation tasks for
each starting model.}
\label{tab:dataset}
\end{table}

\subsection{Coverage and Splits}

Each cell holds two executions, see Table~\ref{tab:dataset}. Cell averages provide the training labels. Tasks, not
cells, are the units of evaluation.

BFCL uses task-grouped cross-validation as it does not possess a separate validation or test split. \taur{} and MultiWOZ have disjoint
training and evaluation tasks. BFCL and MultiWOZ support all three starts on the
same evaluation populations. In \taur{}, incomplete downward handoff coverage limits
the 35B and 397B starts to 17 tasks, compared with 41 at 8B. We report those smaller
populations with particular caution, \ie all \taur{} comparisons remain descriptive
rather than establishing cross-dataset effects. We will release the dataset.

\subsection{TwinRouterBench}
\label{sec:datasets-trb}

TwinRouterBench~\cite{yang2026twinrouterbench} is an agentic routing benchmark
with two parts. Its static bank has 970 labelled steps from 520 instances, each one agent trajectory. Of these instances, 81 have more than one labelled step, and 40, the SWE subset, are software engineering tasks from SWE-bench Verified~\cite{jimenez2024swe}. Each step has a verified benchmark label, from which we build our forward targets,
see Section~\ref{sec:method-supervision}. On the bank, we hold out whole instances
and score each model choice against these labels, not against new executions.
The benchmark scores the routers based on four metrics. RowPass counts steps
given at least the labelled tier, RowExact counts steps given exactly that tier, and
TrajPass counts steps whose whole instance passes. CostSave is the share of the
always-strongest bill saved under the benchmark's cost model, which bills the whole
prompt as a new cache write at every tier switch, and it can be negative because a
failed instance is charged a rerun on the strongest tier. Combined is the mean of
the four and is not live task success.
The benchmark's live track runs a router on 100 held-out instances. The benchmark uses a
different model pool, success measure, and cost model, so its scores are not
comparable with those on our dataset.

%% file: sections/5_results.tex
\section{Results}
\label{sec:results}

All routers use offline full-information supervision and causal inputs,
see Section~\ref{sec:method-routers}. Evaluation populations are given in
Table~\ref{tab:dataset}. Unless stated otherwise, results are at the 35B start, the
only start that permits handoff in both directions. Results at the other starts
are in Appendix~\ref{app:codex-secondary}.

\subsection{Later Handoff}

\begin{figure}[!t]
\centering
\includegraphics[width=\columnwidth]{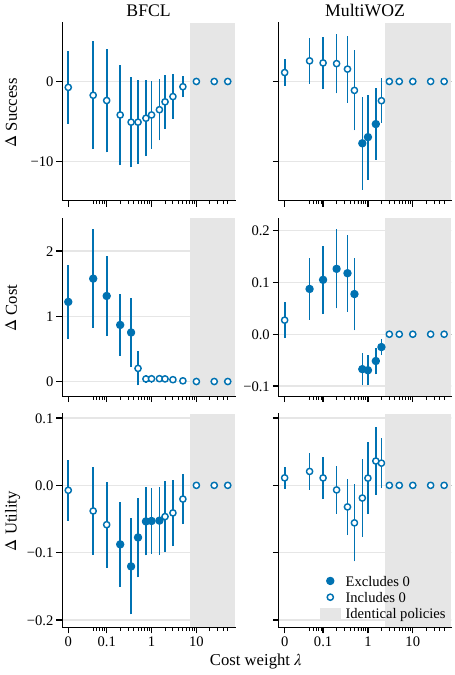}
\caption{Effect of allowing later handoff for the direct router at the 35B start.
Each panel shows later handoff minus initial selection, so values above zero mean
higher success, cost, or utility with later handoff.
Bars are simultaneous bands that cover all cost weights and metrics at once, at
97.5\% per dataset and 95\% over both datasets, and filled markers show bands that
exclude 0. Success differences
are in percentage points and cost differences in $10^{-3}$\,\$ per task.}
\label{fig:handoff-effects}
\end{figure}

No improvement over initial selection is established for the direct router at
the 35B start on BFCL or MultiWOZ at any tested cost weight, see
Figure~\ref{fig:handoff-effects}.
On BFCL, later handoff adds cost: at $\lambda\in[0,0.35]$ it spends 0.75 to
$1.6\times10^{-3}$\,\$ more per task with no established gain in success, and at
$\lambda\in[0.2,1.5]$ it lowers utility. On MultiWOZ, it spends 0.08 to
$0.13\times10^{-3}$\,\$ more per task at $\lambda\in[0.05,0.5]$, and at
$\lambda\in[0.75,1.5]$, where only 7 to 17\% of dialogues succeed, it saves 0.05 to
$0.07\times10^{-3}$\,\$ per task but loses 5.3 to 7.7 points of success. No utility
difference is established at any cost weight on MultiWOZ.
Appendix~\ref{app:codex-secondary} reports sensitivity checks for fold splits, tie
rules, cost forecasts, prefix identity, and waiting rules.

The losses of later handoff come from handoffs that the router ranks poorly. After
the first step, the rank correlation between its predicted and realised success
gain from handing off is about 0.07 on both datasets, where 0
means no rank association. On BFCL at $\lambda=0.2$, the router hands off later to 8B
98 times across its six training seeds, on 49 tasks, predicting higher success than staying despite a
higher forecast cost. These handoffs succeed 44.4\% of the time at
$4.75\times10^{-3}$\,\$ per task, whereas staying on 35B would have succeeded
77.0\% of the time at $4.47\times10^{-3}$\,\$. On MultiWOZ, the router systematically under-predicts success from
staying. With such a weak ranking, the comparison has little power to detect
value in timing.

The factored router and SR-KNN behave in the same way: at $\lambda=0.2$ on BFCL,
their later handoffs cost 0.78 and $0.57\times10^{-3}$\,\$ more per task and lower
utility, with no established change in success, see
Table~\ref{tab:policy-differences}. The 8B and 397B starts permit only upward and
downward handoffs, respectively, see Figure~\ref{fig:codex-frontiers-all}. At the
8B start on BFCL, later handoff adds 4.7 points of success at $\lambda=1$ at almost
no extra cost, which we state as a hypothesis for a new test. On the 47 tasks where
it hands off later, an immediate upward handoff, at $k=0$, would have done better,
with 72.1\% against 32.1\% success for 3.83 against $3.31\times10^{-3}$\,\$ per
task. At the 397B start and $\lambda=0.2$, later handoffs lower both success and cost on
both datasets, see Table~\ref{tab:policy-differences}.

\subsection{Router Comparison}

\begin{figure*}[tp]
\centering
\includegraphics[width=\textwidth]{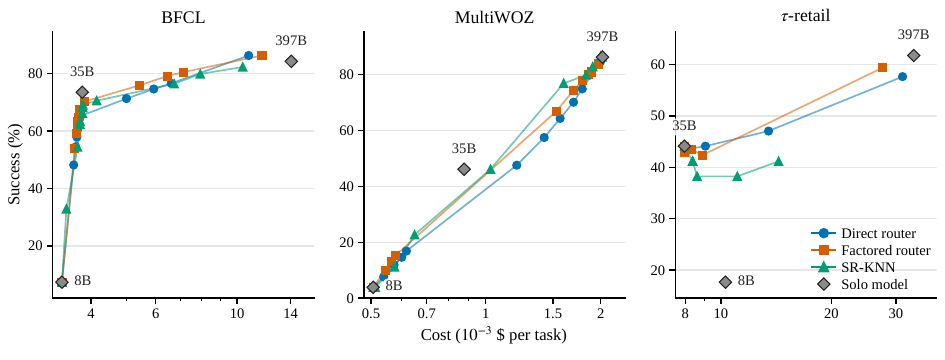}
\caption{Initial selection at the 35B start. Each marker is a router's operating
point at one of the 15 cost weights, and lines join the points in order of
increasing $\lambda$. \taur{} has 17 evaluation tasks at this start.}
\label{fig:initial-selection}
\end{figure*}

Because later handoff adds no established benefit, we compare the routers under
initial selection, see Figure~\ref{fig:initial-selection}. By unadjusted 95\% intervals of the paired utility
difference, the factored router has higher utility than the direct router on
MultiWOZ at 6 of the 15 cost weights, with no established decrease at any weight.
On BFCL, the factored router has higher utility than the direct router at only 2 of
the 15 weights. On MultiWOZ, SR-KNN has higher utility than the factored router at 3
of the 15 weights and lower utility at none, and its highest success is also higher,
86.8\% against 83.8\%. On BFCL and \taur{}, the factored router reaches the higher
success, 86.4\% against 82.4\% and 59.4\% against 41.2\%, at higher cost, 11.69
against 10.36 and 27.64 against $14.36\times10^{-3}$\,\$ per task. A
likely reason for the factored router's gain is its training target: it learns from
success curves made monotone across handoff steps, the same projection that
improves the direct router in Section~\ref{sec:results-supervision}. On \taur{},
which has only 17 evaluation tasks at this start, the comparison is descriptive.

At one low cost weight, $\lambda=0.05$, the factored router has higher utility than
the training-selected fixed policy on BFCL, by 0.055 $[0.015,0.098]$, with 8.5
points more success for $2.30\times10^{-3}$\,\$ more per task.
At $\lambda=0$, both the factored and the direct router reach slightly higher success
than always using 397B at 17 to 23\% lower cost, although the success intervals
include zero. At
$\lambda\in[0.2,2]$ on BFCL, the fixed policy, which stays on 35B, beats the
factored router, and at $\lambda\in[0.2,3]$ it beats SR-KNN. On MultiWOZ, the
factored router establishes no utility improvement over the fixed policy at any weight.

\subsection{Supervision Targets}
\label{sec:results-supervision}

\input{sections/tables/label_effects}

\begin{table*}[t]
\centering\small
\setlength{\tabcolsep}{6pt}
\begin{tabular}{@{}lrrrrr@{}}
\toprule
Policy & RowPass & RowExact & TrajPass & CostSave & Combined \\
\midrule
\multicolumn{6}{@{}l}{\textit{Full bank}} \\
\textcolor{black!55}{Always strongest} & \textcolor{black!55}{100.0} & \textcolor{black!55}{17.5} & \textcolor{black!55}{100.0} & \textcolor{black!55}{0.0} & \textcolor{black!55}{54.4} \\
Burst-aware, benchmark labels & 95.1 & 20.1 & 90.3 & 54.3 & 64.9 \\
Burst-aware, forward targets & \textbf{99.3} & \underline{26.1} & \textbf{98.1} & \textbf{60.6} & \underline{71.0} \\
SR-KNN & \underline{95.7} & \textbf{76.8} & \underline{93.0} & \underline{55.4} & \textbf{80.2} \\
\addlinespace
\multicolumn{6}{@{}l}{\textit{SWE}} \\
\textcolor{black!55}{Always strongest} & \textcolor{black!55}{100.0} & \textcolor{black!55}{50.0} & \textcolor{black!55}{100.0} & \textcolor{black!55}{0.0} & \textcolor{black!55}{62.5} \\
Burst-aware, benchmark labels & 88.4 & 41.4 & 76.8 & \textminus1.2 & 51.3 \\
Burst-aware, forward targets & \textbf{99.7} & \textbf{50.6} & \underline{97.3} & \textbf{2.1} & \textbf{62.4} \\
SR-KNN & \underline{98.2} & \underline{49.7} & \textbf{97.6} & \underline{\textminus0.1} & \underline{61.4} \\
\bottomrule
\end{tabular}
\caption{TwinRouterBench static scores (\%). Our burst-aware router is trained on
the benchmark labels or on our forward targets, see Section~\ref{sec:method-supervision}.
Benchmark labels are TwinRouterBench's tier labels for each step, and forward targets
give each step the strongest tier needed from that step to the end of the instance.
The full bank has 520 instances with 970 labelled steps, and SWE is its subset of
40 SWE-bench Verified instances. Always strongest, in grey, is a reference. Among the
three router rows, the highest value in each column is in bold and the second highest
is underlined.}
\label{tab:trb-supervision}
\end{table*}

On our dataset, projected labels improve the direct router under initial
selection, see Table~\ref{tab:label-effects}. At the same $\lambda$, the two label
types can reach different costs, so we compare each projected operating point with
the success that raw labels reach at the same cost. At the same cost, projected labels
reach 0.2 to 4.1 points more success in all eight cases, with the largest gain on
BFCL at $\lambda=0.5$. Utility, which accounts for cost at each weight, rises in 5
of the 8 cases, with no established decrease, see Table~\ref{tab:label-sweep-bfcl} for all 15
weights. On MultiWOZ, most of the gain at the same weight comes from a more
expensive operating point. Nevertheless, projected labels improve the router
consistently, with more success at the same cost in all eight cases and no established
decrease in utility at any tested weight. Projected labels also raise later handoff, by 2.5 points of success on BFCL and
11.5 on MultiWOZ at $\lambda=0.2$, for 0.18 and $0.21\times10^{-3}$\,\$ more per
task. Later handoff therefore still adds no established benefit. On BFCL it lowers
utility by 0.100 $[0.060,0.141]$ under projected labels, and on MultiWOZ no utility
difference is established.

On TwinRouterBench, our forward targets improve the burst-aware router over the
benchmark labels, see Table~\ref{tab:trb-supervision}. RowExact rewards matching the
benchmark's tier label at each step, whereas CostSave also accounts for the cost
of changing tiers, see Section~\ref{sec:datasets-trb}. Following labels that rise
and fall can therefore improve RowExact while incurring additional cache-write
costs. Both burst-aware variants use the same one-handoff rule. Forward targets
change when and to which tier the router hands off, but they do not introduce the
one-handoff restriction. We report all four components alongside Combined to
distinguish tier matching, successful execution under the benchmark labels, and
cost saving.

On the full bank, forward targets raise RowPass from 95.1 to 99.3, TrajPass from
90.3 to 98.1, and CostSave by 6.3 points. On SWE, the router fails 1 instance
instead of 10, with TrajPass higher by 20.5 points and CostSave by 3.3 points,
although its 95\% interval, $[-4.1,8.3]$, includes zero. SR-KNN's higher full-bank
Combined comes only from RowExact, and on the full bank the forward-target router is
higher on RowPass, TrajPass, and CostSave. The comparison changes three training
targets together, not one label. On the live track, the router trained with
forward targets resolves 79 of 100 instances at a router cost of \$0.59 per
instance, see Appendix~\ref{app:trb-live}. On both our dataset and TwinRouterBench,
changing only the
supervision targets, with the learner, features, and decision rule fixed, improves
the trade-off between success and cost.

%% file: sections/tables/label_effects.tex
\begin{table*}[t]
\centering\small
\setlength{\tabcolsep}{5pt}
\begin{tabular}{@{}r!{\color{black!25}\vrule}rr!{\color{black!25}\vrule}rr!{\color{black!25}\vrule}r@{\;[}r@{,\,}r@{]\hspace{1.4em}}r@{\;[}r@{,\,}r@{]\hspace{1.4em}}c@{}}
\toprule
& \multicolumn{2}{c!{\color{black!25}\vrule}}{Projected labels} & \multicolumn{2}{c!{\color{black!25}\vrule}}{Raw labels} & \multicolumn{7}{c}{Projected $-$ Raw} \\
\cmidrule(l){6-12}
$\lambda$ & Success & Cost & Success & Cost & \multicolumn{3}{c}{$\Delta$ Success} & \multicolumn{3}{c}{$\Delta$ Utility} & $\Delta$ Success at same cost \\
& (\%) & ($10^{-3}$\,\$) & (\%) & ($10^{-3}$\,\$) & \multicolumn{3}{c}{(\%)} & \multicolumn{3}{c}{} & (\%) \\
\midrule
\multicolumn{12}{@{}l}{\textit{BFCL}} \\
0 & 86.6 & 12.59 & 86.4 & 10.75 & 0.2 & \textminus2.6 & 3.3 & 0.002 & \textminus0.026 & 0.033 & 0.2 \\
0.2 & 73.7 & 4.93 & 71.3 & 4.99 & 2.4 & \textminus0.1 & 5.0 & \textbf{0.027} & 0.003 & 0.051 & 2.7 \\
0.5 & 69.0 & 3.74 & 64.0 & 3.71 & 5.1 & 2.3 & 8.0 & \textbf{0.047} & 0.021 & 0.075 & 4.1 \\
1 & 64.9 & 3.70 & 62.1 & 3.70 & 2.8 & 0.5 & 5.1 & \textbf{0.028} & 0.008 & 0.050 & 2.9 \\
\addlinespace
\multicolumn{12}{@{}l}{\textit{MultiWOZ}} \\
0 & 86.3 & 2.02 & 84.7 & 2.00 & 1.6 & 0.6 & 2.8 & \textbf{0.016} & 0.006 & 0.028 & 1.6 \\
0.2 & 75.4 & 1.75 & 64.4 & 1.57 & 11.0 & 7.0 & 15.1 & \textbf{0.067} & 0.038 & 0.097 & 2.2 \\
0.5 & 56.5 & 1.36 & 47.6 & 1.20 & 8.9 & 4.4 & 13.5 & 0.001 & \textminus0.033 & 0.036 & 1.9 \\
1 & 15.4 & 0.60 & 14.6 & 0.60 & 0.8 & \textminus1.4 & 3.0 & 0.007 & \textminus0.011 & 0.027 & 0.7 \\
\bottomrule
\end{tabular}
\caption{Direct router at the 35B start under initial selection, trained on projected or raw labels, see Section~\ref{sec:method-supervision}, and scored on the same raw outcomes. Success in \%, success differences in percentage points, cost in $10^{-3}$\,\$ per task. $\Delta$ Success and $\Delta$ Utility compare the two label types at the same $\lambda$, with unadjusted 95\% intervals, and bold utility differences have intervals that exclude zero. $\Delta$ Success at same cost compares each projected operating point with the Pareto front of the raw-label operating points over all 15 weights. Linear interpolation between front points gives the raw-label success at the projected point's cost, and beyond the most expensive front point the front keeps its highest success.}
\label{tab:label-effects}
\end{table*}

%% file: sections/6_discussion.tex
\section{Discussion}
\label{sec:discussion}

\paragraph{Initial selection.}
At the 35B start, allowing a later handoff adds no established benefit for the
direct router, the factored router, or SR-KNN, and on BFCL it adds cost. The routers' results are therefore
decided mostly by the model they select at the first step. This result depends on
our model pool and on our costs, which are mostly provider bills that include
prompt-cache discounts, while a handoff to another model starts without that cache. The Qwen
experiments do not test transfer across model families. The result also reflects
routers whose ranking of handing off against staying is close to chance, so it is
not evidence that timing has no value. Because our dataset measures the outcome of
a handoff at every collected step, better predictors can be tested on it directly.

\paragraph{Fixed policies.}
A fixed policy chosen on training data is hard to beat, and how much a router can
gain depends on the model pool. On BFCL, 35B solves 73.5\% of tasks at
$3.78\times10^{-3}$\,\$ per task, against 84.3\% at $14.05\times10^{-3}$\,\$ for
397B. At $\lambda=0$, both routers
reach slightly higher success than always using 397B at 17 to 23\% lower cost,
although the success intervals include zero, and at $\lambda=0.05$ the factored
router also has higher utility. On MultiWOZ, 35B solves 46.2\% of dialogues at
$0.88\times10^{-3}$\,\$ per task, against 86.3\% at $2.02\times10^{-3}$\,\$ for
397B, and the routers stay close to 397B in both success and
cost. In all three datasets, 8B fails most tasks, so routing to it saves little. This
suggests that a pool whose models are closer in capability but differ in price would
leave more room for routing to improve the trade-off between success and cost. Router evaluations
should report the training-selected fixed policy next to the router.

\paragraph{Supervision targets.}
The clearest gains in our study come from what the routers learn to predict, not
from when they may hand off. On our dataset, making training curves monotone gives
the direct router 0.2 to 4.1 points more success at the same cost in all eight
reported cases, and raises its utility at 7 of the 15 cost weights on BFCL and at 4
on MultiWOZ, with no decrease established at any tested weight. The factored router,
which learns from monotone curves, has higher utility than the direct router at 6 of
the 15 weights on MultiWOZ. On TwinRouterBench, forward targets raise the burst-aware
router's TrajPass by 7.8 points and its CostSave by 6.3 points on the full static
bank, and on SWE they reduce failed instances from 10 to 1. A router trained with
these targets resolves 79 of 100 live instances at a router cost of \$0.59 per
instance. On our dataset, projected labels improve both policies by similar amounts.

%% file: sections/a_prediction_live_policy.tex
\section{Prediction and Live Policy Details}
\label{app:codex-instruments}

\subsection{Dataset Router Training}

The direct router is fitted once per dataset, fold, and initialisation seed,
pooling all three starting models. In its weighted cross-entropy loss, when a cell
supplies several views, including solo anchors and
stay rows at several handoff steps, those views share its total weight of two
within each fitted objective. Zero-success cells remain in raw-label training.

All routers use frozen \texttt{all-MiniLM-L6-v2} embeddings with normalised
384-dimensional outputs~\cite{wang2020minilm,reimers2019sentence}. The direct router
concatenates the instruction embedding of the first 2,000 characters, 11 history
statistics, 32 history-embedding PCA components fitted on training rows, and 11
model, step, and price features. History statistics describe actions, results,
errors, history length, user turns, and the size and parse health of the carried
state. The static features encode the model pair, direction, current and candidate
steps, their difference, and a price proxy at the median training task length.
In all reported analyses, the candidate step is the current step. At step zero, the start,
direction, and price features are zeroed.

Neural heads use standard scaling, ReLU activations, and Adam~\cite{kingma2015adam},
with learning rate $10^{-3}$, $L_2$ coefficient $10^{-4}$, and at most 700
iterations. BFCL uses partition seed 0, and seeds 1 and 2 test sensitivity to the
split.

For the direct router's label comparison, projection is fitted separately by
training task and ordered model pair. It includes the receiving-model solo
anchor and leaves stay labels raw. Zero-anchor curves remain in training without
an anchor cap. All other training and evaluation settings are held fixed.

\subsection{Factored Router and SR-KNN Baseline}

For the factorisation in Section~\ref{sec:method-routers}, curves with
$d\le10^{-6}$ are excluded from fitting the timing head $g$ but remain in fitting the
difficulty head $d$ and in evaluation. Reconstructed training targets are capped at the solo anchor. The
direct router's projected labels have no such cap.
Both heads use the same neural settings as above. The difficulty head sees the
instruction and the stronger-model indicator on BFCL, or the receiving-model indicator within
each fixed start on MultiWOZ and \taur{}. The timing head sees the instruction,
11 history statistics, starting- and receiving-model indicators, and handoff step ($k/16$ on BFCL and \taur{},
$k/12$ on MultiWOZ). Predicted timing profiles are projected using only steps
observed so far, with the upward handoff profile anchored at one at step zero.

SR-KNN draws neighbours from the same ordered model pair, with one pool entry per
training task and measured cell. Similarity is the dot product of concatenated
normalised instruction and history embeddings. SR-KNN averages the raw labels of the
five most similar cells, with similarity weights floored at $10^{-6}$.

\subsection{Bank Relabelling and Live Handoff}

TwinRouterBench verifies each tier substitution under its search
procedure~\cite{yang2026twinrouterbench}. Its step labels do not measure the outcome
of a permanent handoff, see Section~\ref{sec:method-supervision}.

The classifiers use scikit-learn's histogram gradient
boosting~\cite{friedman2001gbm,ke2017lightgbm}, with seed 0, learning rate 0.1, and
at most 100 iterations and 31 leaves. History excludes system messages, limits
messages to 300 characters and tool-call arguments to 160, and retains the final
4,000 characters. Role-delimited segments are limited to 1,500 characters, with tool
outputs attached to the preceding segment. Features concatenate the embedding of the first 2,000 characters of the first user
message, the embeddings of the last four segments with zero left-padding, the mean of
all segment embeddings, and a scenario indicator. With forward
targets, the burst classifier has three classes: a burst at the current step, a burst
within the next three labelled steps, or neither. Its predicted probability of the
last class is $p_{\mathrm{clear}}$. With benchmark labels, it has two classes, a burst
at the current step or not, and $p_{\mathrm{clear}}$ is the probability of no burst.

Static evaluation uses five outer folds grouped by instance. Within each training
fold, three inner folds supply calibration scores and select
$p_{\min}\in\{0,0.95,0.975,0.98,0.99\}$ and $\beta\in\{0.5,1,2\}$ by the highest
Combined score among settings with positive CostSave. Evaluation uses the original
benchmark labels.

Static cost forecasts use $S_t=(\rho_3-\rho_t)\hat W_{\mathrm{rem}}$ and
$F_t=\beta\rho_3\hat W_{\mathrm{inst}}$, where $\rho_t$ is the tier's input price per
token, and $\hat W_{\mathrm{rem}}$ and $\hat W_{\mathrm{inst}}$ are the forecast prompt
volumes of the remaining steps and of the whole instance. Each prompt-volume forecast multiplies the current history length by a
training mean ratio for remaining or total volume, grouped by scenario and step bin:
0, 1--4, 5--14, 15--39, or 40 and above.

For live use, the router is refitted on the full static bank with three
instance-grouped calibration folds, the software-engineering scenario indicator,
and $p_{\min}=0.95$. Forecasts assume an instance length of 32 steps, at least five remaining steps, 500
output tokens per step, and one prompt token per four characters, using locked input
and output prices without cache discounts. $S_t$ is the remaining cost difference
between tiers 3 and $t$, and $F_t$ is the forecast tier-$t$ spend plus the \$0.60
unresolved-instance penalty. Every eligible tier has positive $V_t$, so the floor $p_{\min}$
controls handoff timing and the handoff score $V_t$ selects the tier. If no handoff has occurred,
spending above \$1.50 or a step count above 70 triggers a permanent handoff to tier 2.
Appendix~\ref{app:trb-live} reports the live evaluation.

%% file: sections/b_execution_inventory.tex
\section{Data Collection and Provenance}
\label{app:codex-data}

BFCL's solo inventory includes 95 screening-only tasks outside the 102 tasks with
handoff curves used for evaluation.

\subsection{Collection and Serving Conditions}

BFCL and MultiWOZ have both handoff directions for every selected task, using the
steps in Table~\ref{tab:handoff-steps}. In \taur{}, all 41 evaluation tasks have
upward handoff curves, but only 17 have downward handoff curves. The 28 upward and 29 downward training
tasks are disjoint, leaving no training task with both directions at the 35B start.
Nine downward training cells are missing: seven had only one valid execution and two
were never collected. MultiWOZ training and evaluation partitions both come from the
MultiWOZ 2.4 validation set.

Qwen3 8B, Qwen3.5 35B-A3B, and Qwen3.5 397B-A17B were served through OpenRouter and
pinned to the providers Alibaba, Parasail, and Chutes, respectively. Provider changes, fallbacks, and campaign-specific exceptions are
recorded by campaign and, where available, by call.

\subsection{Replication and Exclusions}

The dataset retains two executions per cell, selected by source-file order and record
index. It excludes 837 earlier-protocol \taur{} training cells: 768 replayed a
recorded prefix and billed only the continuation, and 69 used a shared recorded
opening. These were not separately collected full episodes.

%% file: sections/c_evaluation_details.tex
\section{Evaluation and Statistical Details}
\label{app:codex-evaluation}

\subsection{Decision Inputs and Canonical Trajectories}

Timing features use the current step index, never its fraction of the final
episode length. Other inputs follow the causal restrictions in
Section~\ref{sec:method-routers}.

Ties between alternative receiving models go to the lower forecast cost and then
to the smaller model. A threshold that allows handoff on ties and rules that give
waiting an explicit value are descriptive sensitivities, see
Appendix~\ref{app:codex-secondary}, not replacements for the primary rule.

One canonical trajectory per task and starting model supplies every candidate's
prefix at a decision point. The canonical trajectory is the starting model's solo
execution with the lowest replicate identifier. If none exists, it is a rollout with
the starting model's prefix and the latest collected handoff step, breaking ties by
fixed dataset keys, not outcomes. The trajectory is selected once, never stitched or replaced,
and its termination is respected.

\subsection{Cost Forecasts and Cost Weights}

Cost forecasts use training cells only. Initial selection uses each model's mean
solo cost. Later handoffs use means by starting model, receiving model, and step.
Staying uses solo means over training tasks whose canonical trajectory reaches that
step.

Later-step means require five distinct training tasks. Unsupported steps use
interpolation within the supported range, the nearest supported step outside it, or
the pooled ordered-pair mean if no step is supported. Initial solo means are exempt
from this requirement. The primary forecasts have no monotonicity constraint.
Reported costs are the selected cells' billed execution costs.

The 15 cost weights are $0$, $0.05$, $0.1$, $0.2$, $0.35$, $0.5$, $0.75$, $1$,
$1.5$, $2$, $3$, $5$, $10$, $25$, and $50$. The normaliser $\bar C_a$ is computed
for each fold, and at the 35B start it is $3.78\times10^{-3}$\,\$ per task on BFCL,
averaged over folds, and $0.87\times10^{-3}$\,\$ on MultiWOZ. No cost weight is
selected on evaluation outcomes.

\subsection{Evaluation Populations and Uncertainty}
\label{app:evaluation-populations}

Evaluation tasks are excluded from training at every starting model. Handoffs are
considered only before the median training task length: 4 steps on BFCL, 8 on
MultiWOZ, and 8, 10, and 12 on \taur{} at the 8B, 35B, and 397B starts. This removes
36 of 276 BFCL and 56 of 771 MultiWOZ decision points.

The data were examined before the protocol was frozen, and screening outcomes also
supply the solo outcomes at $k=0$. Evaluation therefore does not use an untouched
sample, and independent solo outcomes were not re-collected. The effect of changing stay-observation
weights was not tested, so the router's under-prediction of staying on MultiWOZ remains a diagnostic
rather than an established causal explanation.

Task-bootstrap resamples~\cite{efron1979bootstrap} are shared across policies and
cost weights. Primary simultaneous bands~\cite{montielolea2019bands} use the maximum
standardised bootstrap deviation across success, cost, utility, and all 15 weights,
with error rate 0.025 per dataset for 95\% coverage across BFCL and MultiWOZ.
Intervals outside the primary family, including all \taur{} intervals, are
unadjusted 95\% percentile intervals from the same resamples.

\subsection{Fixed-Policy Comparison}

For each dataset, start, fold, and cost weight, the fixed policy maximises mean
training utility over the three solos and handoff schedules observed on at least
80\% of training tasks. At the \taur{} 35B start, only solos qualify. Ties favour
lower mean billed cost, earlier handoff, and then the smaller model. Evaluation
outcomes are never imputed.

For comparisons with the fixed policy, initial selection uses a standardised
opening, taken from one fixed execution of each task, and solo-mean cost forecasts. On BFCL and
MultiWOZ, this produces the same direct-router initial choices as the later-handoff
comparison. On \taur{}, the two can differ.

%% file: sections/d_secondary_comparisons.tex
\section{Additional Results and Sensitivities}
\label{app:codex-secondary}

\subsection{Policy and Supervision Comparisons}

\begin{figure*}[tp]
\centering
\includegraphics[width=\textwidth]{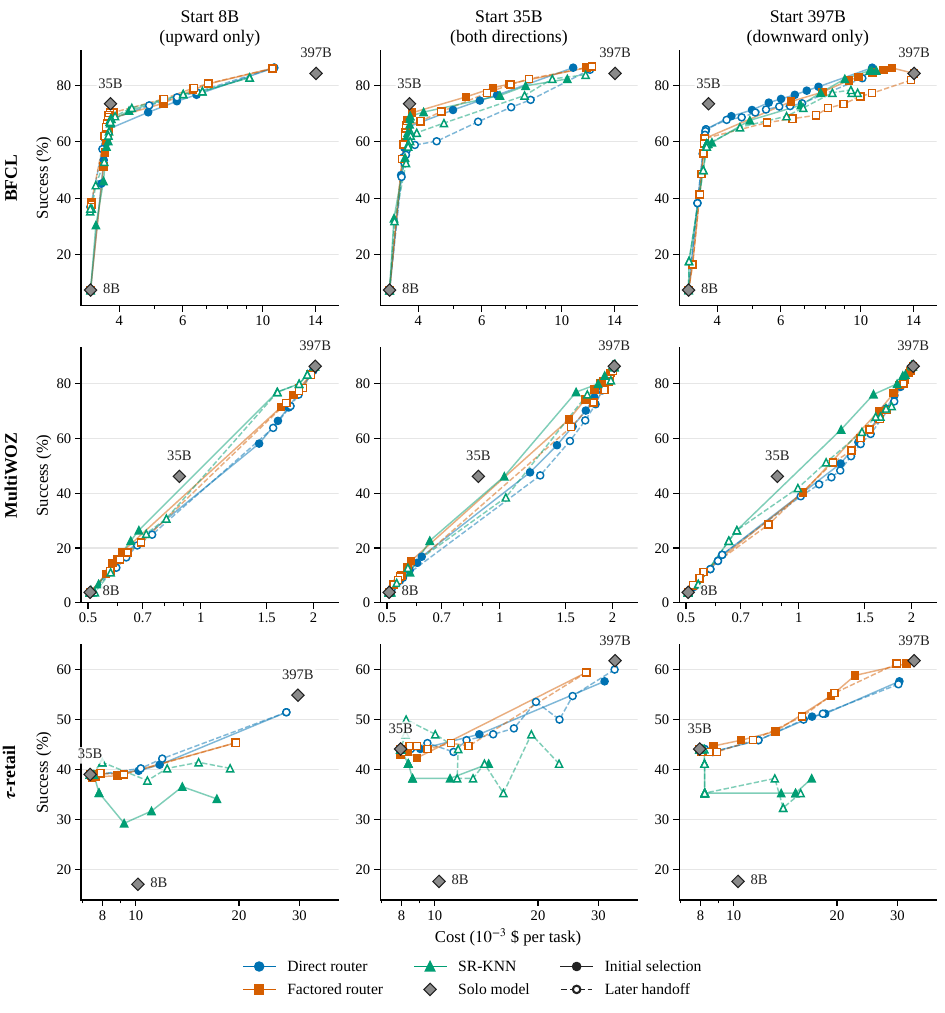}
\caption{Initial selection (solid, filled) and later handoff (dashed, open) at every
starting model over the 15 cost weights. \taur{} has 17 evaluation tasks at the
35B and 397B starts and 41 at the 8B start, see Table~\ref{tab:dataset}.}
\label{fig:codex-frontiers-all}
\end{figure*}

Figure~\ref{fig:codex-frontiers-all} shows initial selection and later handoff for
the two routers and SR-KNN at every starting model, and Table~\ref{tab:policy-differences}
gives their paired differences at $\lambda=0.2$. At the 35B and 397B starts,
\taur{} has 17 evaluation tasks. Table~\ref{tab:sweep-bfcl} gives the values
plotted in Figure~\ref{fig:handoff-effects}.
Table~\ref{tab:label-sweep-bfcl} gives the raw-against-projected label comparison of
Table~\ref{tab:label-effects} at every cost weight.

\input{sections/tables/current_results}

\subsection{Sensitivity Checks}

At BFCL's 35B start, waiting variants improve on strict later handoff at some cost
weights. The direct waiting rule establishes no improvement over initial selection,
and it adds supervision for future handoff schedules, so that comparison does not
isolate the effect of waiting. The factored router's waiting rule is compared only
with strict later handoff.

BFCL partition seeds 1 and 2 preserve the signs of the checked contrasts on the
same 102 tasks. Allowing handoff on ties changes no direct- or factored-router
decisions, and one SR-KNN decision per seed on MultiWOZ at the 397B start. At the
35B start, alternative cost tables change no direct-router decisions on BFCL and
two on MultiWOZ. A monotone cost forecast lowers the success gain of later handoff
at BFCL's 8B start at $\lambda=1$ from 4.7 to 3.3 points. Restricting evaluation to
identical prefixes leaves the BFCL utility difference at $\lambda=0.35$ unchanged
at $-0.120$ on 58 tasks. On 91 MultiWOZ dialogues, the small positive utility
differences at $\lambda=1.5$ and 2 shrink to $+0.008$.

\subsection{Live TwinRouterBench Evaluation}
\label{app:trb-live}

On the live track's 100 held-out instances, the burst-aware router trained with
forward targets on the full static bank resolves 79 of 100 instances at \$0.591 in router cost and \$0.717 in penalty-inclusive leaderboard
bill per instance. Published reference runs resolve 74 instances for always strongest
and 75 for SR-KNN, at agent costs of \$0.547 and \$0.556 per instance, or
bills of \$0.703 and \$0.706 with the same penalty.

The live pool maps tiers 0--3 to DeepSeek V3.2, MiniMax M2.7, Gemini 3 Flash
Preview, and Claude Opus 4.6, respectively. This is the execution pool, distinct
from the static bank's tier annotations.

%% file: sections/tables/current_results.tex
\begin{table*}[p]
\centering\footnotesize
\setlength{\tabcolsep}{4pt}
\begin{tabular}{@{}llr@{\;[}r@{,\,}r@{]\hspace{2\tabcolsep}}r@{\;[}r@{,\,}r@{]\hspace{2\tabcolsep}}r@{\;[}r@{,\,}r@{]}}
\toprule
Start & Router & \multicolumn{3}{c}{$\Delta$ Success (\%)} & \multicolumn{3}{c}{$\Delta$ Cost ($10^{-3}$\,\$)} & \multicolumn{3}{c@{}}{$\Delta$ Utility} \\
\midrule
\multicolumn{11}{@{}l}{\textit{BFCL}} \\
8B & Direct & 2.5 & 1.1 & 4.0 & 0.03 & \textminus0.01 & 0.07 & 0.023 & 0.009 & 0.038 \\
 & Factored & 1.6 & 0.4 & 3.1 & 0.02 & \textminus0.01 & 0.06 & 0.015 & 0.002 & 0.031 \\
 & SR-KNN & 1.0 & 0.0 & 2.9 & 0.08 & \textminus0.01 & 0.22 & 0.005 & \textminus0.002 & 0.018 \\
35B & Direct & \textminus4.2 & \textminus8.0 & \textminus0.7 & 0.87 & 0.60 & 1.16 & \textminus0.088 & \textminus0.127 & \textminus0.052 \\
 & Factored & 1.4 & \textminus0.6 & 3.6 & 0.78 & 0.46 & 1.15 & \textminus0.027 & \textminus0.047 & \textminus0.009 \\
 & SR-KNN & \textminus3.9 & \textminus10.3 & 2.0 & 0.57 & 0.25 & 0.95 & \textminus0.069 & \textminus0.130 & \textminus0.014 \\
397B & Direct & \textminus4.2 & \textminus6.3 & \textminus2.2 & \textminus0.62 & \textminus0.82 & \textminus0.44 & \textminus0.033 & \textminus0.052 & \textminus0.015 \\
 & Factored & \textminus11.3 & \textminus15.8 & \textminus6.9 & \textminus1.83 & \textminus2.25 & \textminus1.44 & \textminus0.087 & \textminus0.128 & \textminus0.046 \\
 & SR-KNN & \textminus4.9 & \textminus9.8 & \textminus0.5 & \textminus0.69 & \textminus1.10 & \textminus0.32 & \textminus0.039 & \textminus0.087 & 0.004 \\
\addlinespace
\multicolumn{11}{@{}l}{\textit{MultiWOZ}} \\
8B & Direct & 5.7 & 3.5 & 8.2 & 0.13 & 0.09 & 0.17 & 0.006 & \textminus0.014 & 0.028 \\
 & Factored & 1.5 & 0.3 & 2.9 & 0.05 & 0.03 & 0.08 & \textminus0.006 & \textminus0.018 & 0.004 \\
 & SR-KNN & 0.0 & 0.0 & 0.0 & 0.00 & 0.00 & 0.01 & \textminus0.002 & \textminus0.005 & 0.000 \\
35B & Direct & 2.2 & 0.1 & 4.5 & 0.13 & 0.08 & 0.17 & \textminus0.007 & \textminus0.028 & 0.014 \\
 & Factored & \textminus0.1 & \textminus1.8 & 1.6 & 0.11 & 0.06 & 0.17 & \textminus0.026 & \textminus0.046 & \textminus0.008 \\
 & SR-KNN & 1.3 & \textminus1.7 & 4.3 & 0.15 & 0.07 & 0.25 & \textminus0.021 & \textminus0.063 & 0.014 \\
397B & Direct & \textminus17.4 & \textminus22.5 & \textminus12.5 & \textminus0.33 & \textminus0.40 & \textminus0.27 & \textminus0.140 & \textminus0.188 & \textminus0.093 \\
 & Factored & \textminus18.0 & \textminus24.2 & \textminus11.9 & \textminus0.37 & \textminus0.44 & \textminus0.30 & \textminus0.143 & \textminus0.202 & \textminus0.084 \\
 & SR-KNN & \textminus15.4 & \textminus21.8 & \textminus9.4 & \textminus0.27 & \textminus0.36 & \textminus0.19 & \textminus0.126 & \textminus0.187 & \textminus0.069 \\
\addlinespace
\multicolumn{11}{@{}l}{\textit{\taur{}}} \\
8B & Direct & 0.0 & 0.0 & 0.0 & 0.03 & 0.00 & 0.10 & \textminus0.001 & \textminus0.003 & 0.000 \\
 & Factored & 0.7 & 0.0 & 2.0 & 0.04 & \textminus0.05 & 0.16 & 0.006 & \textminus0.003 & 0.019 \\
 & SR-KNN & 8.5 & 1.2 & 17.1 & 1.56 & 0.04 & 3.44 & 0.041 & \textminus0.040 & 0.135 \\
35B & Direct & 10.0 & \textminus0.6 & 21.2 & 11.63 & 7.25 & 16.41 & \textminus0.300 & \textminus0.443 & \textminus0.171 \\
 & Factored & 1.2 & \textminus3.5 & 7.1 & 1.58 & 0.40 & 2.99 & \textminus0.043 & \textminus0.114 & 0.020 \\
 & SR-KNN & 2.9 & 0.0 & 8.8 & 5.35 & 1.76 & 9.48 & \textminus0.155 & \textminus0.295 & \textminus0.039 \\
397B & Direct & 0.0 & 0.0 & 0.0 & 0.00 & 0.00 & 0.00 & 0.000 & 0.000 & 0.000 \\
 & Factored & \textminus1.8 & \textminus3.5 & 0.0 & \textminus1.86 & \textminus4.48 & \textminus0.20 & \textminus0.006 & \textminus0.027 & 0.015 \\
 & SR-KNN & 2.9 & 0.0 & 8.8 & \textminus0.58 & \textminus1.73 & 0.00 & 0.033 & 0.000 & 0.099 \\
\bottomrule
\end{tabular}
\caption{Later handoff minus initial selection for each router at $\lambda=0.2$, with success differences in percentage points and cost differences in $10^{-3}$\,\$ per task. Intervals are unadjusted 95\%, and rows of zeros mean that the two policies make identical choices. \taur{} has 41 evaluation tasks at the 8B start and 17 at the 35B and 397B starts, see Table~\ref{tab:dataset}.}\label{tab:policy-differences}
\end{table*}

\begin{table*}[p]
\centering\footnotesize
\setlength{\tabcolsep}{4pt}
\begin{tabular}{@{}rr@{\;[}r@{,\,}r@{]\hspace{2\tabcolsep}}r@{\;[}r@{,\,}r@{]\hspace{2\tabcolsep}}r@{\;[}r@{,\,}r@{]}}
\toprule
$\lambda$ & \multicolumn{3}{c}{$\Delta$ Success (\%)} & \multicolumn{3}{c}{$\Delta$ Cost ($10^{-3}$\,\$)} & \multicolumn{3}{c@{}}{$\Delta$ Utility} \\
\midrule
\multicolumn{10}{@{}l}{\textit{BFCL}} \\
0 & \textminus0.7 & \textminus5.2 & 3.8 & 1.22 & 0.65 & 1.79 & \textminus0.007 & \textminus0.052 & 0.038 \\
0.05 & \textminus1.7 & \textminus8.4 & 5.0 & 1.58 & 0.82 & 2.33 & \textminus0.038 & \textminus0.103 & 0.027 \\
0.1 & \textminus2.4 & \textminus8.8 & 4.1 & 1.31 & 0.70 & 1.93 & \textminus0.058 & \textminus0.122 & 0.006 \\
0.2 & \textminus4.2 & \textminus10.4 & 2.1 & 0.87 & 0.39 & 1.34 & \textminus0.088 & \textminus0.151 & \textminus0.024 \\
0.35 & \textminus5.1 & \textminus10.7 & 0.6 & 0.75 & 0.22 & 1.28 & \textminus0.120 & \textminus0.192 & \textminus0.049 \\
0.5 & \textminus5.1 & \textminus10.3 & 0.1 & 0.20 & \textminus0.06 & 0.46 & \textminus0.077 & \textminus0.137 & \textminus0.018 \\
0.75 & \textminus4.6 & \textminus9.3 & 0.1 & 0.04 & \textminus0.03 & 0.10 & \textminus0.054 & \textminus0.104 & \textminus0.003 \\
1 & \textminus4.2 & \textminus8.4 & 0.0 & 0.04 & \textminus0.02 & 0.10 & \textminus0.053 & \textminus0.101 & \textminus0.004 \\
1.5 & \textminus3.5 & \textminus7.3 & 0.3 & 0.04 & \textminus0.01 & 0.10 & \textminus0.052 & \textminus0.103 & \textminus0.002 \\
2 & \textminus2.5 & \textminus5.9 & 0.8 & 0.04 & \textminus0.01 & 0.09 & \textminus0.046 & \textminus0.099 & 0.006 \\
3 & \textminus1.9 & \textminus4.6 & 0.9 & 0.03 & \textminus0.01 & 0.06 & \textminus0.041 & \textminus0.090 & 0.008 \\
5 & \textminus0.7 & \textminus1.9 & 0.6 & 0.01 & \textminus0.01 & 0.03 & \textminus0.020 & \textminus0.057 & 0.017 \\
10--50 & \multicolumn{9}{l}{identical policies (all differences 0)} \\
\addlinespace
\multicolumn{10}{@{}l}{\textit{MultiWOZ}} \\
0 & 1.1 & \textminus0.5 & 2.7 & 0.03 & \textminus0.01 & 0.06 & 0.011 & \textminus0.005 & 0.027 \\
0.05 & 2.6 & \textminus0.3 & 5.4 & 0.09 & 0.03 & 0.15 & 0.021 & \textminus0.007 & 0.048 \\
0.1 & 2.3 & \textminus1.0 & 5.6 & 0.11 & 0.04 & 0.17 & 0.011 & \textminus0.020 & 0.042 \\
0.2 & 2.2 & \textminus1.5 & 5.9 & 0.13 & 0.05 & 0.20 & \textminus0.007 & \textminus0.042 & 0.028 \\
0.35 & 1.5 & \textminus2.6 & 5.7 & 0.12 & 0.04 & 0.19 & \textminus0.032 & \textminus0.074 & 0.010 \\
0.5 & \textminus1.1 & \textminus6.1 & 3.9 & 0.08 & 0.01 & 0.15 & \textminus0.056 & \textminus0.113 & 0.001 \\
0.75 & \textminus7.7 & \textminus13.5 & \textminus1.9 & \textminus0.07 & \textminus0.10 & \textminus0.04 & \textminus0.019 & \textminus0.077 & 0.039 \\
1 & \textminus6.9 & \textminus12.2 & \textminus1.6 & \textminus0.07 & \textminus0.10 & \textminus0.04 & 0.011 & \textminus0.043 & 0.064 \\
1.5 & \textminus5.3 & \textminus9.7 & \textminus0.9 & \textminus0.05 & \textminus0.08 & \textminus0.03 & 0.036 & \textminus0.014 & 0.086 \\
2 & \textminus2.4 & \textminus5.2 & 0.4 & \textminus0.02 & \textminus0.04 & \textminus0.01 & 0.033 & \textminus0.004 & 0.070 \\
3--50 & \multicolumn{9}{l}{identical policies (all differences 0)} \\
\bottomrule
\end{tabular}
\caption{Later handoff minus initial selection for the direct router at the 35B start, at every cost weight, with success differences in percentage points and cost differences in $10^{-3}$\,\$ per task. Intervals are the primary simultaneous bands of Figure~\ref{fig:handoff-effects}, covering all cost weights and metrics at 97.5\% per dataset and 95\% over both datasets.}\label{tab:sweep-bfcl}\label{tab:sweep-multiwoz}
\end{table*}

\begin{table*}[p]
\centering\footnotesize
\setlength{\tabcolsep}{4pt}
\begin{tabular}{@{}rr@{\;[}r@{,\,}r@{]\hspace{2\tabcolsep}}r@{\;[}r@{,\,}r@{]\hspace{2\tabcolsep}}r@{\;[}r@{,\,}r@{]}}
\toprule
$\lambda$ & \multicolumn{3}{c}{$\Delta$ Success (\%)} & \multicolumn{3}{c}{$\Delta$ Cost ($10^{-3}$\,\$)} & \multicolumn{3}{c@{}}{$\Delta$ Utility} \\
\midrule
\multicolumn{10}{@{}l}{\textit{BFCL}} \\
0 & 0.2 & \textminus2.6 & 3.3 & 1.83 & 1.34 & 2.36 & 0.002 & \textminus0.026 & 0.033 \\
0.05 & 4.7 & 2.3 & 7.3 & 1.68 & 1.13 & 2.27 & 0.025 & 0.002 & 0.049 \\
0.1 & 4.0 & 1.4 & 6.7 & 0.85 & 0.39 & 1.34 & 0.017 & \textminus0.006 & 0.041 \\
0.2 & 2.4 & \textminus0.1 & 5.0 & \textminus0.06 & \textminus0.42 & 0.27 & 0.027 & 0.003 & 0.051 \\
0.35 & 4.4 & 1.8 & 7.3 & \textminus0.01 & \textminus0.05 & 0.03 & 0.045 & 0.019 & 0.072 \\
0.5 & 5.1 & 2.3 & 8.0 & 0.02 & \textminus0.01 & 0.06 & 0.047 & 0.021 & 0.075 \\
0.75 & 3.7 & 1.1 & 6.4 & 0.02 & \textminus0.02 & 0.05 & 0.033 & 0.010 & 0.058 \\
1 & 2.8 & 0.5 & 5.1 & 0.00 & \textminus0.03 & 0.03 & 0.028 & 0.008 & 0.050 \\
1.5 & 1.9 & \textminus0.6 & 4.2 & \textminus0.01 & \textminus0.04 & 0.03 & 0.021 & \textminus0.000 & 0.042 \\
2 & 2.9 & 0.2 & 5.4 & 0.00 & \textminus0.03 & 0.04 & 0.027 & 0.005 & 0.049 \\
3 & 2.1 & \textminus1.4 & 5.6 & 0.01 & \textminus0.03 & 0.05 & 0.015 & \textminus0.023 & 0.053 \\
5 & 0.0 & \textminus3.5 & 3.4 & 0.00 & \textminus0.04 & 0.04 & \textminus0.003 & \textminus0.055 & 0.050 \\
10--50 & \multicolumn{9}{l}{identical policies (all differences 0)} \\
\addlinespace
\multicolumn{10}{@{}l}{\textit{MultiWOZ}} \\
0 & 1.6 & 0.6 & 2.8 & 0.02 & 0.01 & 0.04 & 0.016 & 0.006 & 0.028 \\
0.05 & 6.7 & 3.3 & 10.1 & 0.11 & 0.05 & 0.16 & 0.061 & 0.030 & 0.092 \\
0.1 & 9.5 & 5.6 & 13.4 & 0.14 & 0.07 & 0.21 & 0.079 & 0.045 & 0.113 \\
0.2 & 11.0 & 7.0 & 15.1 & 0.19 & 0.12 & 0.26 & 0.067 & 0.038 & 0.097 \\
0.35 & 8.9 & 4.6 & 13.2 & 0.17 & 0.09 & 0.24 & 0.022 & \textminus0.005 & 0.049 \\
0.5 & 8.9 & 4.4 & 13.5 & 0.15 & 0.08 & 0.23 & 0.001 & \textminus0.033 & 0.036 \\
0.75 & 0.9 & \textminus1.5 & 3.4 & 0.01 & \textminus0.01 & 0.02 & 0.001 & \textminus0.021 & 0.024 \\
1 & 0.8 & \textminus1.4 & 3.0 & 0.00 & \textminus0.01 & 0.01 & 0.007 & \textminus0.011 & 0.027 \\
1.5 & \textminus0.3 & \textminus2.4 & 1.8 & 0.00 & \textminus0.01 & 0.01 & \textminus0.003 & \textminus0.021 & 0.017 \\
2 & 1.3 & \textminus0.8 & 3.7 & 0.00 & \textminus0.01 & 0.02 & 0.005 & \textminus0.016 & 0.025 \\
3--50 & \multicolumn{9}{l}{identical policies (all differences 0)} \\
\bottomrule
\end{tabular}
\caption{Projected minus raw labels for the direct router at the 35B start under initial selection, at every cost weight, with success differences in percentage points and cost differences in $10^{-3}$\,\$ per task. Intervals are unadjusted 95\%.}\label{tab:label-sweep-bfcl}\label{tab:label-sweep-multiwoz}
\end{table*}